\documentclass[10pt,twocolumn,letterpaper]{article}

\usepackage{iccv}
\usepackage{times}
\usepackage{epsfig}
\usepackage{graphicx}
\usepackage{amsmath}
\usepackage{amssymb}
\usepackage{algorithm}
\usepackage{algpseudocode}

\usepackage[pagebackref=true,breaklinks=true,letterpaper=true,colorlinks,bookmarks=false]{hyperref}

\iccvfinalcopy 

\def\iccvPaperID{12} 

\ificcvfinal\fi

\begin{document}

\title{SortedAP: Rethinking evaluation metrics for instance segmentation}

\author{Long Chen$^1$ \and Yuli Wu$^1$ \and Johannes Stegmaier$^1$ \and Dorit Merhof$^2$ \\
$^1$ Institute of Imaging \& Computer Vision, RWTH Aachen University, Germany\\
$^2$ Faculty of Informatics and Data Science, University of Regensburg, Germany\\
{\tt\small \{Long.Chen, Yuli.Wu, Stegmaier.Johannes\}@lfb.rwth-aachen.de,} \\
{\tt\small Dorit.Merhof@informatik.uni-regensburg.de}
}

\maketitle
\ificcvfinal\thispagestyle{empty}\fi

\begin{abstract}
	Designing metrics for evaluating instance segmentation revolves around comprehensively considering object detection and segmentation accuracy. However, other important properties, such as sensitivity, continuity, and equality, are overlooked in the current study. In this paper, we reveal that most existing metrics have a limited resolution of segmentation quality. They are only conditionally sensitive to the change of masks or false predictions. For certain metrics, the score can change drastically in a narrow range which could provide a misleading indication of the quality gap between results. Therefore, we propose a new metric called sortedAP, which strictly decreases with both object- and pixel-level imperfections and has an uninterrupted penalization scale over the entire domain. We provide the evaluation toolkit and experiment code at \url{https://www.github.com/looooongChen/sortedAP}.
\end{abstract}

\section{Introduction}

Recently, considerable work has been conducted in instance segmentation due to its wide scope of application~\cite{he2017mask,wang2020solov2,chen2019tensormask,chen2022instance}, such as autonomous driving~\cite{de2017semantic}, medical diagnosis~\cite{kumar2017dataset} and agricultural phenotyping~\cite{scharr2016leaf,chen2022high}. In the field of bioimage computing, segmenting instances of animals~\cite{mazur2020deep}, cells~\cite{edlund2021livecell}, and subcellular structures~\cite{caicedo2019nucleus,glory2007automated} is also common and infrastructural processing for further analysis and study. Instance segmentation not only localizes the object of interest but also delineates the exact boundary, which can be seen as performing object detection and semantic segmentation concurrently.

Correspondingly, a qualified evaluation metric should consider three fundamental types of imperfections: missed ground truth objects (false negative), falsely predicted objects (false positive), and segmentation inaccuracy. Existing metrics all incorporate the three error types above, but are not discussed with respect to properties, including sensitivity, continuity, and equality.

\textbf{Sensitivity.} An ideal metric should be sensitive to all occurrences of imperfections of all types. Any additional errors are supposed to lead monotonically to a worse score, not ignored or obscured by the occurrence of other errors. A metric that monotonically decreases with any errors will enable a more accurate comparison.

\textbf{Continuity.} The penalization scale of a metric should be relatively consistent locally across the score domain. Intuitively, gradually and evenly changing segmentations should correspond to a smoothly changing metric score as well. Abrupt changes are not desired.

\textbf{Equality.} Without any assumed importance of different objects, all objects should have an equal influence on the metric score. A common case of inequality is that the score is biased towards larger objects. Although larger objects may be prioritized in some applications, as a general metric, the metric should treat all objects equally. Analysis with respect to object size can be easily performed by evaluating different size groups using a metric of equal property.

Although all metrics discussed in this paper implement a penalization of false positive, false negative, and segmentation inaccuracy, the majority of metrics, even very widely used ones, such as the mean Average Precision (mAP)~\cite{caicedo2019nucleus}, are only conditionally sensitive to errors. This violates the sensitivity property, as some differences in segmentation results are not reflected in the score. For match-based approaches, such as Average Precision (AP)~\cite{caicedo2019nucleus} and Panoptic Quality (PQ)~\cite{kirillov2019panoptic}, the score will change abruptly at the match threshold. There is actually a paradox in choosing thresholds, which is discussed in Section~\ref{sec:deficiency}.

To address the gap, we propose a new metric called the sorted Average Precision (sortedAP). Unlike mAP~\cite{caicedo2019nucleus}, which queries the AP score at a sequence of fixed intersection over union (IoU) thresholds, sortedAP detects every exact IoU value at which the AP score drops. This is achieved through our proposed \emph{Unique Matching} approach and sorting all possible matches according to the IoU values (Section~\ref{sec:sortedAP}). The Unique Matching method explicitly preserves the one-to-one relationship between two sets of instances. This also allows the use of IoU thresholds smaller than 0.5, or under object overlap, in all match-based metrics.

\section{Related work: A review}

This section provides an overview of proposed evaluation metrics in the literature. We use the notion $\mathcal{G}=\{g_1,g_2, \dots, g_M\}$ and $\mathcal{P}=\{p_1,p_2, \dots, p_N\}$ to represent the set of ground truth and predicted objects in the following context. The capitalized symbols $\mathcal{G}$ and $\mathcal{P}$ can represent a set, or the number of elements in the set, for notation simplicity.

\subsection{Overlap-based metrics}
The Dice coefficient (Dice) and the Intersection over Union (IoU) are the most commonly used metrics to measure the similarity between two binary masks. The IoU, also known as Jaccard Index (JI), is defined as the ratio of the intersection area to the union area between two masks: 
\begin{equation}
	IoU(p,g) = \frac{|p \cap g|}{|p \cup g|}.
\end{equation}

Instead of the union, Dice use accumulated area: 

\begin{equation}
	Dice(p,g) = \frac{2 \cdot |p \cap g|}{|p|+|g|}. 
\end{equation}

Although they have slightly different definitions, both metrics utilize the same fact that the intersection area is maximized when two masks are identical. Furthermore, the two metrics are directly related in values:

\begin{equation}
	Dice(p,g) = \frac{2 \cdot IoU(p,g)}{1+IoU(p,g)}.
\end{equation}

\textbf{Aggregated Jaccard Index (AJI).} The AJI~\cite{kumar2017dataset} extends the Jaccard Index to instance segmentation by accumulating the object-level intersection and union area, which is computed between each ground truth object and the prediction yielding the maximum IoU. The area of predicted objects without any matched ground truth objects is also aggregated to the union area as the penalization to false positives.

\textbf{Symmetric Best Dice (SBD).} SBD~\cite{scharr2016leaf} is based on an asymmetric score Best Dice (BD). For each object in one set, BD finds the maximal Dice with any object in the other set (the reference set) for averaging.

\begin{equation}
BD(\mathcal{P},\mathcal{G})=\frac{1}{N}\sum_{i=1}^{N}\max_{j=1:M}Dice(p_i,g_j),
\end{equation} 

The BD does not fully penalize all errors, since unmatched objects in the reference set are excluded and have no impact on the score. Therefore, the SBD computes BD using both sets under comparison as the reference and takes the worse score as the final score:

\begin{equation}
	SBD(\mathcal{P},\mathcal{G}) = min\{BD(\mathcal{P},\mathcal{G}), BD(\mathcal{G},\mathcal{P})\}.
\end{equation}

\subsection{Match-based metrics}

Another category of metrics is based on object-level detection errors at one or multiple segmentation quality thresholds. A matching criterion $t$, typically an IoU value, is defined as a prerequisite. Each ground truth object searches for a successful match in the predicted objects, or vice versa. Based on the match results, all objects can be grouped into one of the three categories: true positives ($TP_t$), false positives ($FP_t$), and false negatives ($FN_t$).

Fundamentally, the match between predicted objects and ground truth objects should satisfy a one-to-one relationship. This ensures that the number of true positives is equal to the number of ground truth objects that have a successful match. We will discuss how to explicitly maintain this relationship in Section~\ref{sec:unique_match}.

\textbf{Average precision (AP).} The term AP can refer to different evaluation metrics in the literature. For ease of discussion, we refer to them as the P-R AP~\cite{everingham2010pascal} and the point AP~\cite{caicedo2019nucleus}. Despite being based on different perspectives, both metrics are defined in terms of precision and recall:

\begin{equation}
Pre_t = \frac{TP_t}{TP_t+FP_t},\; Rec_t = \frac{TP_t}{TP_t+FN_t}.
\end{equation}

The P-R AP was first proposed for the evaluation of object detection tasks~\cite{everingham2010pascal,lin2014microsoft}. As a summary of the Precision-Recall curve (P-R curve), it evaluates a model from a more comprehensive view by considering the precision performance over the entire recall domain. Although very widely used, the P-R AP suffers from certain deficiencies, as pointed out by recent works. Firstly, the definition requires a confidence score for each prediction, while not all approaches naturally score the outputs.
For example, most bottom-up approaches do not directly deliver object-level confidence scores as most detection-based pipelines do. In terms of discrimination capability, P-R AP does not really distinguish between different shapes of P-R curves~\cite{oksuz2018localization}. The neglect of low-confidence duplicates (hedged prediction) is another important deficiency of P-R AP~\cite{jena2023beyond}.

In comparison, the point AP is oriented towards the end result and corresponds to a point on the P-R curve that achieves a certain precision-recall trade-off. In this case, all predictions are treated equally regardless of scoring. The point AP is formulated as follows:

\begin{equation}
	AP_t = \frac{TP_t}{TP_t+FP_t+FN_t}.
\end{equation}

The point AP relates to the P-R curve according to the following equation:

\begin{equation}
	AP_t = \frac{1}{Pre_t+Rec_t+1}.
\end{equation}

While the P-R AP favors precision improvements at any recall level, the point AP only focuses on the single point of best precision-recall trade-off. From the user's perspective, higher precision in the extreme recall range is of limited practical significance. Therefore, point AP obligates the processing pipeline to screen predictions, including determining the optimal cutoff confidence. In the following context, we refer to the point AP when using the term AP.

\textbf{Mean Average Precision (mAP).} The AP score is based on the matching results under a certain IoU threshold $t$. Segmentation imperfections better than the matching criterion will not be further penalized. Similarly, objects worse than the threshold are viewed as equally bad.

To compensate for the neglect of segmentation imperfections, the mean Average Precision (mAP)~\cite{caicedo2019nucleus} averages a series of AP scores over progressively higher IoU thresholds:

\begin{equation}
	mAP = \frac{1}{N}\sum_{t\in T}\frac{TP_t}{TP_t+FP_t+FN_t},
\end{equation}
where $T=\{t_1, t_2, \dots, t_N \}$. A typical choice for the threshold range is from 0.5 to 0.95, with a step size of 0.05. 

It is worth mentioning that when referring to mAP, it generally means the averaging of multiple AP scores, rather than scores under different matching thresholds specifically. For example, the PASCAL dataset ~\cite{everingham2010pascal} computes P-R AP scores of different semantic classes for averaging. The COCO challenge~\cite{lin2014microsoft} considers both the semantic categories and varying matching thresholds. In this work, we only discuss averaging across matching thresholds, as it is directly relevant to metric design.

\textbf{Panoptic Quality (PQ).} The PQ is defined as the multiplication of the Recognition Quality (RQ) 
\begin{equation}
	RQ = \frac{2 \cdot TP_{t=0.5}}{2 \cdot TP_{t=0.5}+FP_{t=0.5}+FN_{t=0.5}}
\end{equation}
and the Segmentation Quality (SQ)
\begin{equation}
	SQ = \frac{\sum_{(p,g)\in TP_{T=0.5}} IoU(p,g)}{|TP_{t=0.5}|},
\end{equation}
where $(p,g)$ indicates a matched prediction and ground truth pair. The RQ measures the detection accuracy as the AP and they are related as 
\begin{equation}
	RQ = \frac{2 \cdot AP_{t=0.5}}{1+AP_{t=0.5}}.
\end{equation}
The SQ term is basically the mean IoU of all true positive pairs, explicitly modeling the segmentation quality of objects above the match threshold.

\begin{figure*}[h]
	\begin{center}
		\includegraphics{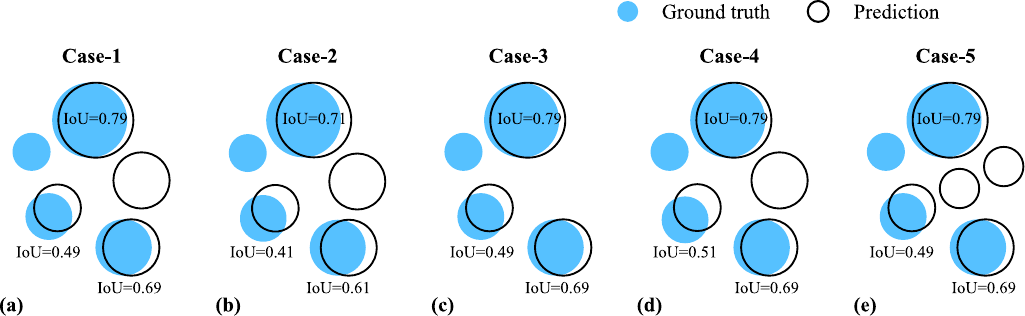}
	\end{center}
	\caption{Examples to illustrate the deficiencies of evaluation metrics. (a) Case-1 is the base example. (b) All IoUs get worse in Case-2, but the mAP score remains unchanged. (c) Case-3 contains one less false positive, but SBD score is the same as Case-1. (d) In Case-4, only one object segmentation improves by 0.02 in IoU, but the PQ score increases by 17.64\%. (e) Two false positives are present in Case-5, while only one exists in Case-1. AJI score penalizes them equally due to the smaller size of objects in Case-5.}
	\label{fig:deficiency}
\end{figure*}

\section{An analysis of deficiencies}
\label{sec:deficiency}

\begin{table}
	\begin{center}
		\begin{tabular}{c|ccccc}
			\hline
			Metrics & AJI & SBD & PQ & mAP & sortedAP \\
			\hline
			Case-1 & \textbf{.5125} & \textbf{.4925} & \textbf{.4229} & \textbf{.3778} & .4261 \\
			Case-2 & .4587 & .4325 & .3771 & \textbf{.3778} & .3839 \\
			Case-3 & .6252 & \textbf{.4925} & .4933 & .4722 & .5283 \\
			Case-4 & .5159 & .4975 & \textbf{.4975} & .4000 & .4288 \\
			Case-5 & \textbf{.5125} & .3940 & .3700 & .3148 & .3572 \\
			\hline
		\end{tabular}
	\end{center}
	\caption{Scores of different metrics for the examples shown in Figure~\ref{fig:deficiency}. In each column, the pair of cases marked in bold demonstrate the deficiency of a metric.}
	\label{tab:deficiency}
\end{table}

\subsection{Sensitivity to errors}
\label{sec:sensitivity}
While existing metrics account for all three types of errors, few of them are sensitive to all occurrences of errors.

\textbf{Exempted error.} SBD takes the worse BD score between using the ground truth and the prediction as the reference. This only considers false positives or false negatives, except the segmentation inaccuracy, respectively. As illustrated in Figure~\ref{fig:deficiency}a and Figure~\ref{fig:deficiency}c, predictions with and without an additional false positive have the same SBD score. Although the false prediction decreases BD$(\mathcal{P},\mathcal{G})$, the impact on SBD is exempted by the lower BD$(\mathcal{G},\mathcal{P})$.

\textbf{Resolution of segmentation difference.} As stated previously, the mAP score reflects the segmentation quality by computing AP scores at varying IoU thresholds, with a certain step size. Despite having a good practical utility with an appropriate step size, mAP is only definitely sensitive to IoU changes greater than the step size. A smaller difference in IoU may or may not result in score changes, depending on whether the change crosses a predefined IoU threshold or not. From Figure~\ref{fig:deficiency}a to Figure~\ref{fig:deficiency}b, all IoUs decrease by 0.08. However, the mAP score remains unchanged in the case of step size 0.1 (Figure~\ref{fig:sortedAP}).

\subsection{Match thresholds and score continuity}
\label{sec:continuity}

Match-based metrics use hard thresholds to determine true and false positives. As a result, objects can abruptly transition from true positives to false positives, even if they are only slightly different in IoU. PQ and mAP introduce a continuous or quasi-continuous measure of the segmentation, but only in the domain above the minimum IoU threshold. A discontinuous change always occurs at the lower IoU threshold. An example is shown in Figure~\ref{fig:deficiency}, where increasing the IoU of only one prediction from 0.49 to 0.51 leads to a PQ change of 17.64\%, from 0.4229 to 0.4975 (Table~\ref{tab:deficiency}).

\textbf{Threshold dilemma.} A discontinuous score is not completely unacceptable. The IoU threshold can be set low enough so that two useful results (away from the low IoU range) will not be assigned drastically different scores. However, a single AP or PQ score reported with a low match threshold becomes less informative. PQ makes the compromise at the IoU of 0.5. The mAP only alleviates the amplitude of abrupt changes by dividing them into multiple levels (Figure~\ref{fig:exp}c and Figure~\ref{fig:exp}f).

\subsection{Equality of object-level errors}

Without specific assumptions, objects should be treated equally. A missed small object is supposed to place the same impact on the score as a larger object. Object segmentation accuracy should also be measured relative to their size, rather than the absolute area. Match-based approaches satisfy this property by constructing the metric using the object counts and object-level IoU. SBD takes the average of object Dice, therefore also area-independent. In contrast, AJI does not have a notion of objects. For instance, the scenario of having two false positives in Figure~\ref{fig:deficiency}e yields the same AJI score as the scenario of having one larger false positive in Figure~\ref{fig:deficiency}a. And accumulating absolute area will also bias the score towards the quality of larger objects.

\section{Sorted Average Precision (sortedAP)}
\label{sec:sortedAP}

\subsection{Unique Matching}
\label{sec:unique_match}

For match-based metrics, each ground truth object can match at most one prediction, and vice versa. This rule ensures that the number of true positives is consistent with the number of ground truth objects that have a successful match. In the greedy match used by mAP and PQ, the one-to-one relationship is implicitly maintained by using matching IoUs larger than 0.5. This is because, under the non-overlapping assumption, no two objects can match with the same object while both having IoUs larger than 0.5~\cite{kirillov2019panoptic}.

We propose using the Hungarian algorithm~\cite{kuhn1955hungarian} to determine true positive matches. This involves the following steps: constructing the cost matrix, padding the cost matrix to square, solving the maximal assignment problem using the Hungarian algorithm, and removing matches of zero cost. The implementation details are depicted in Algorithm~\ref{alg:unique_match}.

The Hungarian matching algorithm not only maintains the one-to-one match relationship but also maximizes the accumulated IoUs of true positive matches. The Unique Matching as a plug-in extension can be applied to both AP and PQ, making them applicable with low match thresholds and object overlap.

\begin{algorithm}
	\caption{Unique Matching}
	\label{alg:unique_match}
	\begin{algorithmic}[h]
		\Require ground truth $\mathcal{G}=\{g_1,g_2, \dots, g_M\}$, prediction $\mathcal{P}=\{p_1,p_2, \dots, p_N\}$, matching threshold $t$
		\Ensure the match matrix $\mathbf{TP} \in \{true, false\}^{N\times M}$
		\State Initialize the cost matrix $\mathbf{Cost} \in R^{N\times M}$
		\For{each prediction $p_i \in \mathcal{P}$ }
		\For{each ground truth object $g_j \in \mathcal{G}$ }
		\If{$IoU(g_i, p_j) > t$}
		\State $\mathbf{Cost}(i,j) = 1-IoU(p_i, g_j)$
		\EndIf
		\EndFor
		\EndFor
		\If{$N>M$}
		\State pad $N-M$ dummy zero columns to $\mathbf{Cost}^{N\times M}$
		\Else
		\State pad $M-N$ dummy zero rows to $\mathbf{Cost}^{N\times M}$
		\EndIf
		\State $\mathbf{TP}^{N\times M} \gets $  run standard Hungarian algorithm, remove dummy rows or columns
		\For{$i$ from 1 to N }
		\For{$j$ from 1 to M }
		\If{$\mathbf{Cost}(i, j) == 0$}
		\State $\mathbf{TP}(i,j) = false$
		\EndIf
		\EndFor
		\EndFor
	\end{algorithmic}
\end{algorithm}

\begin{figure}
	\begin{center}
		\includegraphics{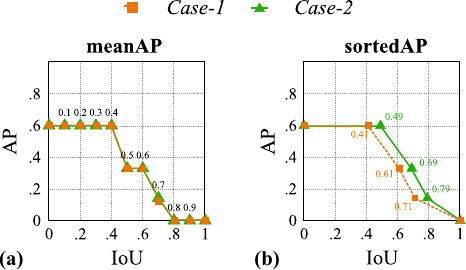}
	\end{center}
	\caption{Computation of meanAP and sortedAP on the Case-1 and Case-2 in Figure~\ref{fig:deficiency}. The mAP estimates the AP curve by querying AP values at fixed IoUs, while sortedAP identifies the exact IoU value where the AP curve drops.}
	\label{fig:sortedAP}
\end{figure}

\subsection{AP scores over the entire IoU domain}

To avoid the drastic score change (Section~\ref{sec:continuity}), we propose to summarize the AP scores over the entire IoU threshold domain as a metric, instead of a single AP score or scores covering only part of the domain. By using our proposed Unique Matching approach, the mAP can be straightforwardly extended to the entire IoU domain, such as using a threshold collection of $\{0.1, 0.2, ..., 0.9\}$. However, querying AP scores at fixed IoU values can ignore small segmentation changes, noted as the limited resolution in Section~\ref{sec:sensitivity}.

We propose sorted Average Precision (sortedAP) as a new metric that is sensitive to all segmentation changes. The concept of sortedAP involves identifying all IoU values at which the AP score drops, instead of querying AP scores at fixed IoUs as the mAP. The AP score can only change at the IoUs of each object where the object transitions from true positive to false positive. Raising the matching threshold from 0 to 1 will turn all matches into non-matches one by one in the ascending order of IoU. In consequence, one non-match will diminish a true positive and introduce a false negative. Considering the sum of true and false positives is constant, we rewrite the AP score as:

\begin{equation}
	AP_t = \frac{TP_t}{TP_t+FP_t+FN_t} = \frac{TP_t}{P+FN_t}.
\end{equation}

We let $TP_0$ and $FN_0$ be true positives and false negatives of the maximal possible match between two sets. This can be obtained by the Unique Matching (Section~\ref{sec:unique_match}) with a tiny but non-zero fuzzy threshold. All possible AP scores can then be computed by:

\begin{equation}
	AP_{t_k} = \frac{TP_0-k}{P+FN_0+k}, \; k=1,2,...,TP_0,
\end{equation}
where $t_k$ is the k-th lowest IoU of all matches. As shown in Figure~\ref{fig:sortedAP}b, any segmentation differences will be reflected by the positions of turning points. The sortedAP is defined as the area under the AP curve and can be computed by Algorithm~\ref{alg:sortedAP}. In the computation of sortedAP, the Unique Matching runs only once, while it has to be performed multiply times for different IoUs in mAP.

\begin{algorithm}
	\caption{Sorted Average Precision}
	\label{alg:sortedAP}
	\begin{algorithmic}[h]
		\Require ground truth $\mathcal{G}=\{g_1,g_2, \dots, g_M\}$, prediction $\mathcal{P}=\{p_1,p_2, \dots, p_N\}$
		\Ensure the sortedAP score $s \in R$
		\State Match: run Unique Matching with a fuzz threshold $1e^{-6}$
		\State Count: true positives $TP_0$ and false negatives $FN_0$
		\State Sort: arrange IoUs of all matches in increasing order $[IoU_1, IoU_2, \dots, IoU_{TP_0}]$
		\State Initialize: $AP_{prev}\gets\frac{TP_0}{P+FN_0}$, $t_{prev} \gets IoU_1$
		\State Initialize: $s\gets t_{prev} \cdot AP_{prev}$
		
		\For{$k$ from 1 to $TP_{0}$}
			\State $AP_k \gets \frac{TP_0-k}{P+FN_0+k}$, $t_k \gets IoU_k$
			\State $s \gets s + \frac{1}{2}\cdot(t_k-t_{prev})\cdot(AP_k + AP_{prev})$
			\State $AP_{prev} \gets AP_k$, $t_{prev} \gets t_k$
		\EndFor
		
		
	\end{algorithmic}
\end{algorithm}

\section{Experiments and results}

\begin{figure*}
	\begin{center}
		\includegraphics{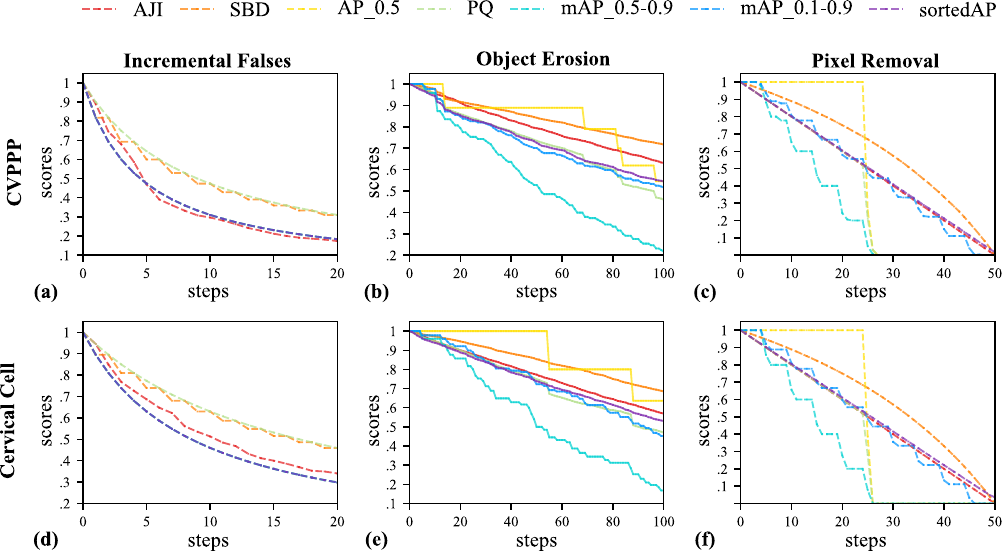}
	\end{center}
	\caption{Comparison of different metric scores on simulated imperfect segmentation results. Three experiments (Incremental Falses, Object Erosion, and Pixel Removal) create increasingly degraded results from the ground truth of real datasets (CVPPP and CervicalCell). Since errors are gradually and evenly introduced, the evaluation score is supposed to smoothly decrease in response. In Figure~\ref{fig:exp}a and Figure~\ref{fig:exp}d, the curve of AP, mAP, and sortedAP are identical, shown in mixed dark blue.}
	\label{fig:exp}
\end{figure*}

We also simulate imperfect results on the basis of ground truth segmentation from real datasets, in order to observe the behavior of different metrics. We choose the CVPPP dataset~\cite{scharr2016leaf} and the CervicalCell dataset~\cite{lu2016evaluation}, containing clustered instances. We perform experiments per image because the effects, such as abrupt changes, will be covered when averaged over a large population. We design three experiments based on the fact that introducing errors gradually and evenly will result in a smooth decrease in the evaluation score.

\textbf{Incremental falses.} This experiment starts with two identical sets of objects and alternately introduces new objects into each set. At each step, we randomly duplicate an object and place it in a position where it does not overlap with any existing objects. This ensures that the newly introduced object is always a false positive or false negative. In our experiment, we add two objects to one set, then switch to the other set and repeat the process. The experiment only concerns detection errors, as objects are either perfectly matched or not matched at all.

\textbf{Object erosion.} At each step, morphological erosion is performed to a random object with a 3$\times$3 structuring element. Consequently, the segmentation quality will steadily deteriorate. But we do not completely remove any objects. Metric scores are reported between the continuously eroded masks and the original set.

\textbf{Pixel removal.} Similar to the object erosion experiment, we construct a sequence of increasingly degraded results by deteriorating the segmentation quality of objects. However, instead of handling one object per step, we randomly remove a fixed portion of pixels from all objects at each step. This process simulates a situation where the segmentation of most objects is at a similar quality level. The deficiencies are more pronounced in this experiment.

The experiments conducted on the CVPPP and CervicalCell datasets yielded similar results. In the incremental false experiment (Figure~\ref{fig:exp}a and Figure~\ref{fig:exp}d), objects are either a perfect match with an IoU of 1 or not matched at all. Thus, segmentation inaccuracy does not play any role. The AP, mAP, and sortedAP all degrade to the same score in this case. All match-based metrics decrease smoothly as expected. In contrast, the AJI fluctuates depending on the size of introduced objects. The SBD score does not decrease in a strictly monotonic manner but instead exhibits periodic plateaus. This is an instance of the error exemption (Section~\ref{sec:sensitivity}). In the alternating introduction of false matches into two sets, errors introduced earlier can obscure subsequent ones.

In the object erosion and pixel removal experiment, the segmentation quality gets worse step by step. The AP and mAP also show plateaus but for a different reason from SBD in the incremental false experiment. This is due to AP's insensitivity to segmentation differences above or below the match threshold. Using multiple thresholds by mAP only improves sensitivity up to the scale of the threshold interval. PQ explicitly considers segmentation quality in the IoU range above the threshold. However, it faces a common issue of abrupt change at the match threshold as AP and mAP. Combining the two factors above, mAP exhibits a step-wise change, which is more noticeable in the pixel removal experiment (Figure~\ref{fig:exp}c and Figure~\ref{fig:exp}f). PQ scores will not be completely flat, but can drastically drop in a narrow IoU range. In comparison, our proposed sortedAP maintains sensitivity and continuity in all cases where other metrics fail.

\section{Conclusion}

In this paper, we have analyzed existing evaluation metrics for instance segmentation from the perspective of sensitivity, continuity, and equality. Although some metrics are widely used in practice, we have found that no metric strictly satisfies all the properties under discussion. To address this gap, we propose the sortedAP, which is sensitive to any small segmentation changes, continuous over the entire IoU domain, and treats objects equally. The proposed Unique Matching approach can also be applied to AP, mAP, and PQ, allowing its use under object overlap and match IoU thresholds smaller than 0.5.



{\small
\bibliographystyle{ieee_fullname}
\bibliography{egbib}
}

\end{document}